\documentclass[conference]{IEEEtran}
\IEEEoverridecommandlockouts
\usepackage{cite}
\usepackage{amsmath,amssymb,amsfonts}
\usepackage{algorithmic}
\usepackage{graphicx}
\makeatletter
\let\IEEEorigmakecaption\@makecaption
\makeatother
\usepackage{subcaption}
\makeatletter
\let\@makecaption\IEEEorigmakecaption
\makeatother
\usepackage{textcomp}
\usepackage{xcolor}
\usepackage{booktabs}
\usepackage{tabularx}
\usepackage{float}
\def\BibTeX{{\rm B\kern-.05em{\sc i\kern-.025em b}\kern-.08em
    T\kern-.1667em\lower.7ex\hbox{E}\kern-.125emX}}

\begin{document}

\title{Navigation with RF Cues: Embodied Perception--Action under Multipath Uncertainty}

\author{
\IEEEauthorblockN{Wenlihan Lu, Tianshun Li, Liuqing Yang, Shijian Gao}
\IEEEauthorblockA{Hong Kong University of Science and Technology (Guangzhou), Guangzhou, China}
}

\maketitle

\begin{abstract}
Smart-factory inspection requires robots to reach connected equipment without a prior map or known target coordinates. Radio-frequency (RF) signals from the target can provide directional cues to complement visual observations when occlusion or poor lighting limits target detection. However, multipath propagation can distort these cues, making it difficult to infer the target's true direction from instantaneous RF measurements. To enable navigation research under these conditions, we first construct a Habitat--Sionna RT benchmark that uses detailed scene geometry and assigned material properties to generate aligned visual and RF observations in response to robot actions. Building on this benchmark, we propose an uncertainty-aware multimodal navigation framework that jointly estimates target direction and its uncertainty from a history of RF, visual, and pose observations. These estimates inform action selection alongside visual context. Experiments in unseen scenes show relative improvements of 18.2\% in success rate (SR) and 11.5\% in success weighted by path length (SPL) over the strongest evaluated baseline.
\end{abstract}


\section{Introduction}
In smart factories, mobile robots may need to approach connected equipment for inspection despite lacking indoor localization information. This poses an embodied navigation task in which the robot must use onboard observations to find and reach a target whose coordinates are unknown. Visual observations support exploration and obstacle avoidance, but machinery and storage racks may obscure the target and limit visual guidance. For equipment with Wi-Fi or Bluetooth Low Energy (BLE) connectivity, radio-frequency (RF) measurements of received signal strength~\cite{wu2017dorfin,lu2025bayesian} and direction of arrival~\cite{kotaru2015spotfi} can provide additional location cues. These measurements exploit signals emitted during normal communication~\cite{muller2023radiosensing,bi2023joint3d} to complement visual observations, without requiring dedicated navigation beacons.

The reliability of these location cues is affected by multipath propagation in indoor environments. Reflections create multiple signal paths between the target and the robot, and a strong reflected component can cause the estimated direction of arrival to deviate from the true target bearing. Interference among paths also causes received signal strength to fluctuate, so stronger reception does not necessarily indicate that the robot is approaching the target. As the robot moves, changes in the relative lengths and strengths of these paths further vary the observed RF cues, making it difficult to navigate consistently toward the target using instantaneous measurements alone.

To mitigate misleading RF cues, mmWave-LSC~\cite{yin2022mmwave} combines path estimation and link-state classification to select between RF directional guidance and visual exploration. PIRL~\cite{li2025digitaltwin} incorporates propagation characteristics into reward design, while PiPRL~\cite{li2025piprl} uses symbolic programs and estimated link states to guide navigation. These methods require detailed channel measurements from mmWave hardware, limiting their direct applicability to Wi-Fi- or BLE-only targets. Moreover, compact receive arrays have limited spatial resolution, making multipath components difficult to distinguish and goal direction uncertain. We therefore jointly estimate goal direction and its uncertainty from multimodal observation history and integrate both with visual context for action selection.

Evaluating navigation under multipath requires an interactive benchmark that supports the perception--action loop of embodied navigation: the robot executes an action and receives visual and RF observations at the resulting pose for the next decision. Existing datasets such as P2SLAM~\cite{arun2022p2slam} and WAIS~\cite{arun2024wais} provide real-world Wi-Fi measurements alongside visual and motion data. However, their fixed trajectories cannot supply observations for arbitrary actions selected by a navigation policy, limiting their use for closed-loop evaluation. The simulation benchmark in mmWave-LSC~\cite{yin2022mmwave} supports such interaction, but its RF simulation relies on simplified floor-plan-based geometry that omits finer scene details. These limitations motivate an interactive RF navigation benchmark that preserves detailed scene geometry and generates aligned visual and RF observations in response to the robot's actions.

To address unreliable RF cues and the lack of interactive benchmarks with detailed scene modeling, we develop a navigation framework and a simulation benchmark. Our framework combines an uncertainty-aware goal predictor with an adapted scene memory transformer~\cite{fang2019scene}. The predictor estimates a target-bearing distribution from a multimodal observation history comprising RF observations (RSSI and the MUSIC spatial spectrum), RGB images, depth images, and pose information, while scene memory integrates it with visual and pose features for action selection. Our Habitat--Sionna RT benchmark uses MP3D scenes~\cite{chang2017matterport3d} to couple embodied navigation~\cite{savva2019habitat} with RF simulation~\cite{hoydis2023sionnart}, preserving detailed geometry and material-dependent propagation effects while generating observations in response to robot actions. Across 1{,}350 episodes in 27 unseen scenes, our method achieves an SR of 0.546 and an SPL of 0.456, exceeding the strongest evaluated baseline by 18.2\% and 11.5\%, respectively.

\section{Problem Statement}
\label{sec:problem}

We consider mobile robot navigation to a stationary goal in an unknown indoor environment. Without a prior map or access to the goal coordinates, the robot must use onboard observations to explore the environment, reach the goal, and stop nearby. Visual and motion observations support exploration and obstacle avoidance, while signals from an RF transmitter at the goal provide additional directional cues that may be distorted by multipath propagation. We specify the task setup and objective below.

\subsection{Task Setup}
\label{sec:task}
\label{sec:rfobs}

Let $\mathcal C$ denote the indoor scene and $\mathcal X_{\mathcal C}$ its navigable space. The goal is located at an unknown position $g\in\mathcal X_{\mathcal C}$. At step $t$, the robot has pose $p_t=(x_t,y_t,\phi_t)$, where $(x_t,y_t)$ is its position and $\phi_t$ its heading. Starting from $p_0$, the robot receives an observation
\begin{equation}
o_t=\left(I_t^{\mathrm{rgb}},I_t^{\mathrm d},u_t,w_t\right),
\label{eq:observation}
\end{equation}
where $I_t^{\mathrm{rgb}}$ and $I_t^{\mathrm d}$ are egocentric RGB and depth images, respectively, $u_t$ denotes odometry relative to the initial pose, and $w_t$ comprises RF cues. The spatial spectrum provides directional cues for navigation; however, reflections and scattering in indoor environments give rise to multipath propagation, causing the estimated arrival directions to potentially deviate from the geometric bearing to the goal. 

At each step, the robot selects an action from
\begin{equation}
\mathcal A=\{\textsc{move\_forward},\textsc{turn\_left},\textsc{turn\_right},\textsc{stop}\}.
\label{eq:action_space}
\end{equation}
We adopt a discrete action configuration~\cite{krantz2020beyond} in which a forward action moves the robot by 0.25\,m, and each left or right turn changes its heading by $15^\circ$. The robot obtains a new observation after each movement and must determine when to stop from its available information.

\subsection{Task Objective}
\label{sec:objective}

The navigation policy selects each action based on all observations received so far and the preceding actions. An episode terminates when the robot selects \textsc{stop} or reaches the maximum number of actions $T_{\max}$.

Let $\tau$ denote the robot's trajectory during a navigation episode, and let $t_{\mathrm{end}}$ denote the final decision step of that episode. An episode is successful if the robot selects \textsc{stop} at a position whose geodesic distance to the goal is less than the success threshold $\epsilon$. The success indicator is defined as
\begin{equation}
\sigma(\tau)\overset{\triangle}{=}\mathrm{I}\!\left[
 a_{t_{\mathrm{end}}}=\textsc{stop}
 \;\land\;
 d_{\mathrm{geo}}(p_{t_{\mathrm{end}}},g)<\epsilon
\right],
\label{eq:success}
\end{equation}
where $\mathrm{I}[\cdot]$ is the indicator function and $d_{\mathrm{geo}}(p_t,g)$ denotes the shortest traversable-path distance from the robot's position at pose $p_t$ to the goal at $g$.

The task requires both successful navigation and efficient travel. Let $L(\tau)$ denote the total distance traveled by the robot and $L^*=d_{\mathrm{geo}}(p_0,g)$ the shortest-path distance from its initial position to the goal. The task-level objective is to maximize the expected SPL:
\begin{equation}
\max_{\pi}\;\mathbb E_{\tau\sim P_{\pi}}\!\left[
\sigma(\tau)\frac{L^*}{\max\{L(\tau),L^*\}}
\right],
\label{eq:spl}
\end{equation}
where $P_{\pi}$ denotes the trajectory distribution induced by the navigation policy and the distribution of scenes, initial poses, and goal locations.

\section{Proposed Solution}
\label{sec:method}

\begin{figure*}[!t]
    \centering
    \includegraphics[width=\textwidth]{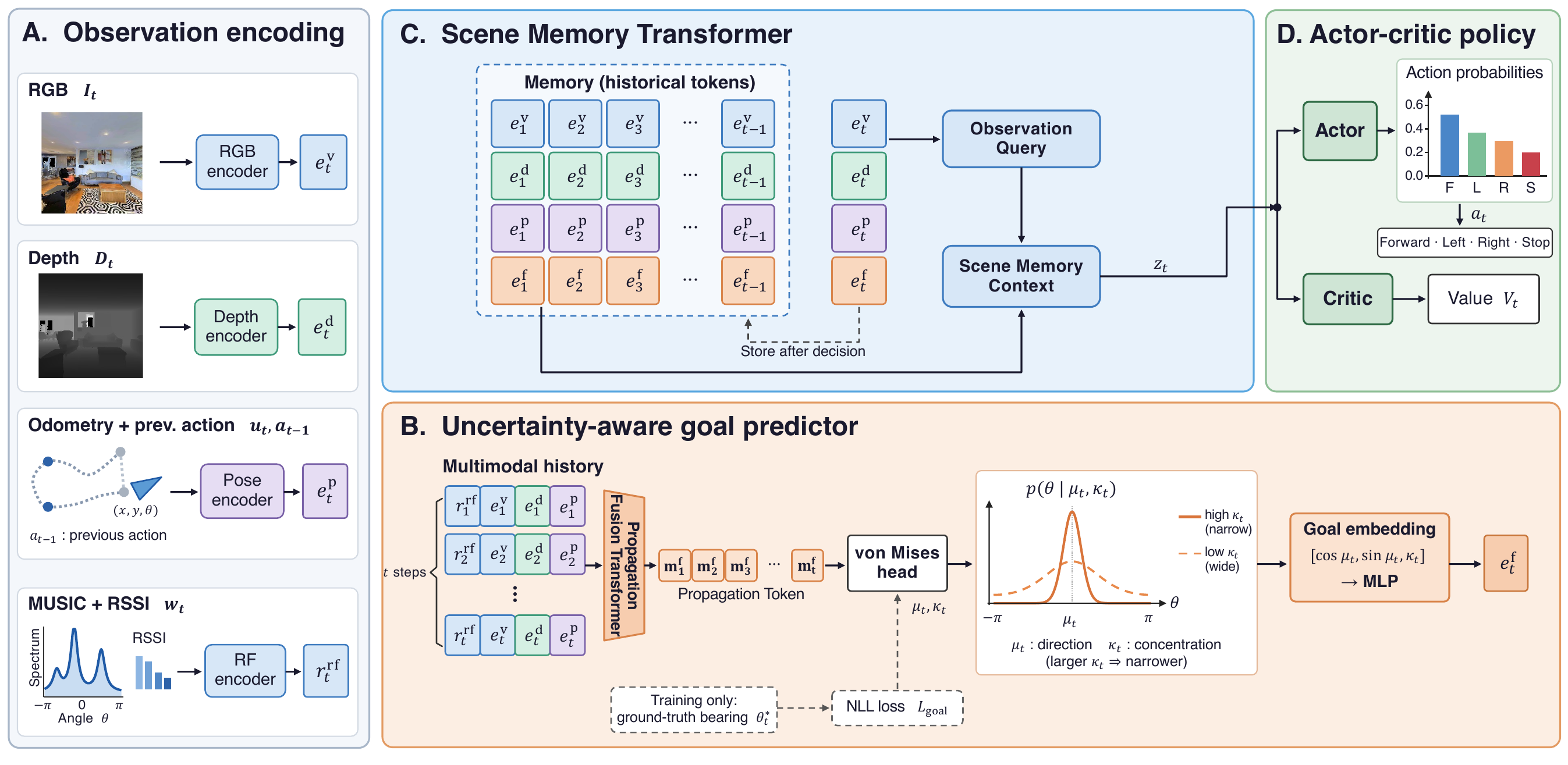}
    \caption{Overview of the proposed framework. The uncertainty-aware goal predictor uses the multimodal observation history to predict a von Mises distribution over the source direction. Its goal tokens are stored alongside RGB, depth, and pose tokens in the scene memory transformer. A query formed from the current observation retrieves historical context through cross-attention for actor critic navigation. Ground-truth direction supervision is used only during training.}
    \label{fig:framework}
\end{figure*}

Fig.~\ref{fig:framework} illustrates the proposed architecture, where RF observations provide goal-direction cues, while visual observations provide scene context for spatial understanding and obstacle avoidance. The uncertainty-aware goal predictor estimates a distribution over the goal direction from the multimodal observation history. The scene memory transformer integrates this distribution with visual and pose features over time to guide navigation.

\subsection{RF and Visual Observation Encoding}
\label{sec:music}

The RF front end uses a forward-facing $1{\times}2$ patch antenna array to obtain RSSI and a MUSIC spatial spectrum~\cite{schmidt1986music}. Let $\mathbf{x}_{t,k}\in\mathbb{C}^{2}$ denote the $k$th array snapshot at time $t$. The sample covariance is
\begin{equation}
\widehat{\mathbf R}_t=\frac{1}{K}\sum_{k=1}^{K}
\mathbf{x}_{t,k}\mathbf{x}_{t,k}^{\mathrm H}.
\label{eq:music_covariance}
\end{equation}
Here, $(\cdot)^{\mathrm H}$ denotes conjugate transpose. We obtain the MUSIC spatial spectrum $P_t(\theta)$ by evaluating the orthogonality between the array steering vectors and the noise subspace estimated from $\widehat{\mathbf R}_t$~\cite{schmidt1986music}. The peak-based direction estimate is
\begin{equation}
\widehat{\theta}_t^{\mathrm{MUSIC}}
=\underset{\theta\in\Theta}{\arg\max}\;P_t(\theta),
\label{eq:music_bearing}
\end{equation}
where $\Theta$ is the set of scanned azimuth angles in the robot frame. Under multipath propagation, the dominant peak may correspond to a reflected arrival rather than the geometric direction to the goal. We therefore include the sampled spectrum, together with RSSI, in the RF observation $w_t$, preserving directional information beyond a single peak for subsequent goal-direction and uncertainty prediction.

We encode the RF observation $w_t$ into a feature $r_t^{\mathrm{rf}}$. Separate encoders produce RGB and depth tokens $e_t^v$ and $e_t^d$, while the pose encoder combines odometry with the previous action to produce $e_t^p$:
\begin{equation}
\begin{aligned}
r_t^{\mathrm{rf}}&=f_{\mathrm{rf}}(w_t), &
 e_t^p&=f_p(u_t,a_{t-1}),\\
e_t^v&=f_v(I_t^{\mathrm{rgb}}), &
 e_t^d&=f_d(I_t^{\mathrm d}).
\end{aligned}
\label{eq:modality_encoding}
\end{equation}
These features provide complementary information for navigation: RF observations provide directional cues, RGB and depth observations provide scene context, and pose features describe robot motion. Their histories are jointly processed by the goal predictor to support source-direction estimation. The goal predictor encodes the estimated direction and its uncertainty into a goal token, which is integrated with visual and pose tokens in the scene memory transformer. This fusion enables the policy to assess goal-directed motion in the context of surrounding obstacles and traversable space, guiding action selection toward the goal.

\subsection{Uncertainty-Aware Goal Predictor}
\label{sec:goal_predictor}

The predictor uses the multimodal observation history to estimate the source bearing and its uncertainty, reducing reliance on MUSIC peaks caused by reflections. A propagation fusion transformer processes the concatenated RF, RGB, depth, and pose feature sequence:
\begin{equation}
m_{1:t}^{f}=F_{\mathrm{prop}}\!\left(
\left\{[r_i^{\mathrm{rf}};e_i^v;e_i^d;e_i^p]\right\}_{i=1}^{t}
\right).
\label{eq:propagation_tokens}
\end{equation}
The current representation $m_t^f$ captures directional evidence and visual scene context across observations and robot motion. We model the target bearing using a von Mises distribution, the circular analogue of the normal distribution. Unlike a linear variable, the bearing is periodic over $[0,2\pi)$, with $0$ and $2\pi$ representing the same direction. The von Mises distribution respects this periodicity and characterizes directional uncertainty through its concentration parameter, making it suitable for bearing prediction. A prediction head maps $m_t^f$ to the mean direction $\mu_t$ and nonnegative concentration $\kappa_t$:
\begin{equation}
p(\theta\mid\mu_t,\kappa_t)=
\frac{\exp\!\left[\kappa_t\cos(\theta-\mu_t)\right]}
{2\pi I_0(\kappa_t)},
\label{eq:von_mises}
\end{equation}
where $\theta$ denotes a candidate bearing in the robot frame and $I_0$ is the modified Bessel function of order zero. Larger $\kappa_t$ indicates a more concentrated prediction. Given the ground-truth bearing $\theta_t^*$, training minimizes the negative log-likelihood:
\begin{equation}
\mathcal L_{\mathrm{goal}}=
\mathbb E_t\!\left[
\log\!\left(2\pi I_0(\kappa_t)\right)
-\kappa_t\cos(\theta_t^*-\mu_t)
\right].
\label{eq:goal_loss}
\end{equation}
Ground-truth bearings are used only for training. Unlike the instantaneous spectral-peak estimate $\widehat{\theta}_t^{\mathrm{MUSIC}}$ in Eq.~\eqref{eq:music_bearing}, $\mu_t$ is predicted from the multimodal observation history. The predicted distribution parameters are encoded into a goal token as $e_t^f=f_{\mathrm{goal}}([\cos\mu_t;\sin\mu_t;\kappa_t])$, where $f_{\mathrm{goal}}$ is an MLP and $[\cdot;\cdot]$ denotes concatenation. The sine and cosine encode the circular bearing, while $\kappa_t$ represents directional concentration. This token carries direction and uncertainty into the scene memory transformer.

\subsection{Memory-Based Navigation Policy}
\label{sec:backbone}
\label{sec:policy_learning}

We adapt the scene memory transformer~\cite{fang2019scene} to RF-informed navigation by augmenting its observation representation with an uncertainty-aware goal token. Specifically, the RGB, depth, and pose tokens $e_t^v$, $e_t^d$, and $e_t^p$ from Eq.~\eqref{eq:modality_encoding} provide visual scene context and robot motion information. The RF feature $r_t^{\mathrm{rf}}$ from the same equation is processed by the goal predictor together with visual and pose history to estimate $\mu_t$ and $\kappa_t$, which are then encoded into the goal token $e_t^f$. The resulting observation representation is $\mathcal E_t=\{e_t^v,e_t^d,e_t^p,e_t^f\}$. By retaining these tokens over time, the scene memory makes historical goal estimates and their uncertainty available alongside visual and motion context for action selection. Historical tokens are processed into the scene memory context:
\begin{equation}
M_{1:t-1}=F_{\mathrm{mem}}\!\left(
\{\mathcal E_i\}_{i=1}^{t-1}\right).
\label{eq:scene_memory}
\end{equation}
An observation query $Q_t$ is formed from the current tokens and retrieves relevant historical features through cross-attention:
\begin{equation}
c_t^{\mathrm{mem}}=\operatorname{softmax}\!\left(
\frac{Q_tK_t^{\top}}{\sqrt{d_k}}\right)V_t^{\mathrm{mem}},
\label{eq:memory_attention}
\end{equation}
where $K_t$ and $V_t^{\mathrm{mem}}$ are the keys and values derived from the scene memory $M_{1:t-1}$, and $d_k$ is the key dimension. Softmax operates over historical tokens. Current tokens are stored after each decision, and memory is reset each episode.

The retrieved context is concatenated with the current observation representation to produce the policy feature $z_t$. The actor maps $z_t$ to a distribution over the four navigation actions, and the critic estimates the expected return:
\begin{equation}
\pi_{\psi}(a_t\mid h_t)=\operatorname{softmax}(f_{\mathrm{actor}}(z_t)),
\end{equation}
\begin{equation}
V_t=f_{\mathrm{critic}}(z_t).
\end{equation}
Here, $\psi$ denotes the policy parameters, and $h_t$ denotes the interaction history comprising all observations received up to time $t$ and the preceding actions. The navigation reward encourages progress toward the goal and successful stopping:
\begin{equation}
\begin{aligned}
r_t={}&\alpha\big(d_{\mathrm{geo},t}-d_{\mathrm{geo},t+1}\big)\\
&+\beta\,\mathrm{I}[D_t=1\land \sigma_t=1]\\
&-\eta\,\mathrm{I}[D_t=1\land \sigma_t=0],
\end{aligned}
\label{eq:navigation_reward}
\end{equation}
where $\alpha$, $\beta$, and $\eta$ are positive coefficients for navigation progress, successful termination, and unsuccessful termination, respectively. We use a larger $\beta$ and a smaller $\eta$ to emphasize successful navigation while applying a modest penalty for failure. The distance is $d_{\mathrm{geo},t}=d_{\mathrm{geo}}(p_t,g)$, $D_t$ indicates episode termination, and $\sigma_t=\mathrm{I}[a_t=\textsc{stop}\land d_{\mathrm{geo},t}<\epsilon]$ is the stepwise success indicator corresponding to Eq.~\eqref{eq:success}. The distance and success signals are used for training and are not policy inputs.

We train the model using proximal policy optimization (PPO)~\cite{schulman2017ppo} to maximize the expected discounted navigation return, together with the auxiliary direction-prediction objective. The joint loss is
\begin{equation}
\mathcal L=\mathcal L_{\mathrm{PPO}}+
 c_v\mathcal L_V-c_e\operatorname{Ent}(\pi_\psi)+
 \lambda_g\mathcal L_{\mathrm{goal}},
\label{eq:joint_objective}
\end{equation}
where $\mathcal L_{\mathrm{PPO}}$ is the clipped policy loss, $\mathcal L_V$ is the value loss, and $\operatorname{Ent}(\pi_\psi)$ is policy entropy. The goal-prediction loss learns a directional representation with uncertainty, while the navigation objective learns how to use that representation and scene memory for action selection.

\begin{figure}[!t]
    \centering
    \includegraphics[width=\columnwidth]{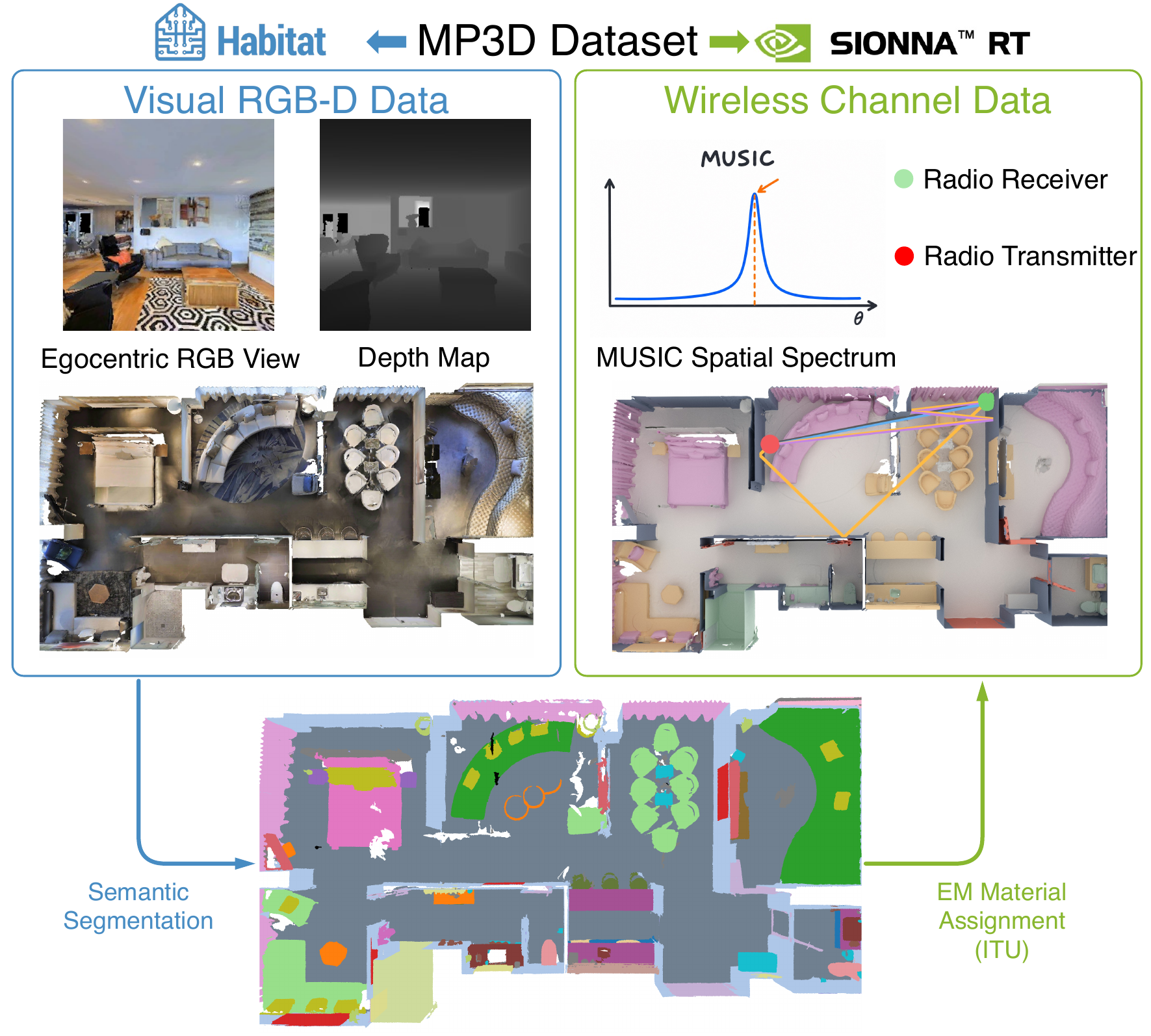}
    \caption{Habitat--Sionna RT benchmark construction from MP3D scenes. Habitat renders egocentric RGB-D observations, while semantically segmented meshes are assigned electromagnetic material properties using ITU models for RF simulation in Sionna RT.}
    \label{fig:benchmark_construction}
\end{figure}

\section{Experiments}
\label{sec:experiments}
\label{sec:benchmark}

\begin{figure}[!t]
    \centering
    \begin{subfigure}[t]{0.49\columnwidth}
        \centering
        \captionsetup{justification=centering,singlelinecheck=false,margin={0.26\linewidth,0pt}}
        \includegraphics[width=\linewidth]{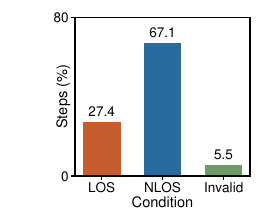}
        \caption{Oracle paths.}
    \end{subfigure}\hfill
    \begin{subfigure}[t]{0.49\columnwidth}
        \centering
        \captionsetup{justification=centering,singlelinecheck=false,margin={0.26\linewidth,0pt}}
        \includegraphics[width=\linewidth]{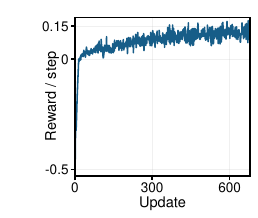}
        \caption{Training reward.}
    \end{subfigure}
    \caption{(a) Distribution of line-of-sight (LOS), NLOS, and invalid observations along shortest-path oracle trajectories. (b) Mean reward per step during training.}
    \label{fig:propagation_reward}

\end{figure}

\begin{table}[!t]
\caption{Training and RF simulation parameters.}
\label{tab:training_parameters}
\centering
\small
\setlength{\tabcolsep}{4pt}
\begin{tabularx}{0.88\columnwidth}{@{}>{\raggedright\arraybackslash}X c@{}}
\toprule
Parameter & Value \\
\midrule
Episode action limit ($T_{\max}$) & 200 \\
Initial geodesic distance & $[8,20]$\,m \\
Success threshold ($\epsilon$) & 1\,m \\
RGB-D resolution & $256\times256$ \\
Reward weights ($\alpha,\beta,\eta$) & $(2.0,10.0,1.0)$ \\
\midrule
Carrier frequency & 2.4\,GHz \\
Receive array & $1\times2$ patch \\
Element spacing & Half wavelength \\
Max. reflection/refraction depth & 5 \\
Receiver thermal noise & Enabled \\
\bottomrule
\end{tabularx}
\end{table}

\begin{table}[!t]
\caption{Main results on 1{,}350 episodes from 27 unseen MP3D scenes.}
\label{tab:main_results}
\centering
\begin{tabularx}{\columnwidth}{@{}>{\raggedright\arraybackslash}X c c c@{}}
\toprule
Method & SR $\uparrow$ & SPL $\uparrow$ & DTG (m) $\downarrow$ \\
\midrule
Random policy & 0.000 & 0.000 & 12.512 \\
RSSI-gradient controller & 0.002 & 0.001 & 10.129 \\
MUSIC-DOA controller & 0.074 & 0.069 & 8.360 \\
mmWave-LSC~\cite{yin2022mmwave} (adapted) & 0.155 & 0.116 & 8.173 \\
PiPRL~\cite{li2025piprl} (adapted) & 0.328 & 0.264 & 7.030 \\
SAVN-CE~\cite{zeng2026savnce} & 0.462 & 0.409 & 5.810 \\
\midrule
Ours & 0.546 & 0.456 & 4.176 \\
\bottomrule
\end{tabularx}
\end{table}

\begin{figure*}[!t]
    \centering
    \includegraphics[width=\textwidth]{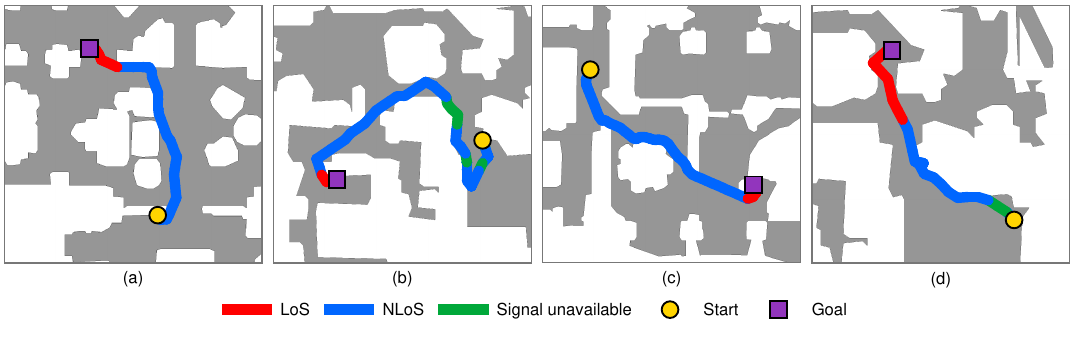}
    \caption{Qualitative trajectories of the proposed method. Colored route segments indicate LOS, NLOS, and unavailable RF signals; markers denote the start and goal.}
    \label{fig:qualitative_episode}
\end{figure*}

\subsection{Benchmark and Experimental Setup}
\label{sec:exp_setup}

Experiments use 90 Matterport3D (MP3D) scenes \cite{chang2017matterport3d} with paired Habitat-navigable meshes and Sionna RT assets. A fixed scene-level split assigns 63 scenes to training and 27 to evaluation without overlap. Habitat Pathfinder samples navigable start and goal locations at geodesic distances of $[8,20]$\,m for training. Evaluation uses 50 fixed episodes per held-out scene at the same distance range, yielding 1{,}350 episodes. The success threshold is $\epsilon=1$\,m, episodes are capped at $T_{\max}=200$ actions, and RGB-D observations are rendered at $256{\times}256$ resolution.

As illustrated in Fig.~\ref{fig:benchmark_construction}, the benchmark uses MP3D scene geometry to generate paired visual and RF observations through Habitat and Sionna RT. Habitat renders egocentric RGB-D images. For RF simulation, we partition the scene meshes according to their semantic annotations, assign material categories to the resulting regions, and specify their electromagnetic properties using ITU material models at the simulation frequency. A fixed coordinate transform aligns the robot poses and transmitter positions between the two simulators. At each navigation step, Sionna RT simulates multipath propagation using the settings in Table~\ref{tab:training_parameters}, providing RF observations corresponding to the robot's current pose.

To characterize propagation conditions, we analyze shortest-path oracle trajectories from each start position to the goal. Along these trajectories, an RF observation is invalid when no propagation path is resolved; otherwise, it is LOS if an unobstructed zero-interaction path exists and NLOS if it does not. Along the oracle trajectories, 27.4\% of observations are LOS, 67.1\% are NLOS, and 5.5\% are invalid (Fig.~\ref{fig:propagation_reward}(a)). Thus, indirect propagation dominates even along the shortest paths.

All methods share the scene split, action space, visual encoders, training observations, and environment-interaction budget whenever their original designs permit. We report SR, SPL, and final distance to goal (DTG), defined as $d_{\mathrm{geo},t_{\mathrm{end}}+1}$. Higher SR/SPL and lower DTG indicate better performance.

\subsection{Comparison Methods and Results}
\label{sec:baselines}
\label{sec:main_results}

We compare the proposed framework with the following baselines:
\begin{itemize}
    \item \textbf{Random policy:} Selects navigation actions randomly.
    \item \textbf{RSSI-gradient controller:} Uses changes in received signal strength to guide navigation toward stronger reception.
    \item \textbf{MUSIC-DOA controller:} Uses the peak direction of the MUSIC spatial spectrum as the estimated target bearing.
    \item \textbf{mmWave-LSC}~\cite{yin2022mmwave}: Combines path estimation and link-state classification to select between directional guidance and visual exploration, adapted to the RF observations available in our setup.
    \item \textbf{PiPRL}~\cite{li2025piprl}: Uses symbolic programs and estimated link states to guide navigation, adapted to the RF observations available in our setup.
    \item \textbf{SAVN-CE}~\cite{zeng2026savnce}: Serves as a memory-based navigation baseline with a deterministic goal representation.
\end{itemize}
All adapted learned baselines are retrained on the same training split. Table~\ref{tab:main_results} reports performance on the held-out evaluation episodes.

The RSSI-gradient and MUSIC-DOA controllers rarely succeed, achieving SRs of 0.002 and 0.074, respectively. These results suggest that relying solely on RF cues makes successful navigation difficult, underscoring the importance of incorporating visual observations. The adapted learned methods perform better, with PiPRL achieving an SR of 0.328 and an SPL of 0.264. SAVN-CE is the strongest baseline, reaching an SR of 0.462 and an SPL of 0.409.

The proposed method achieves the highest SR and SPL, at 0.546 and 0.456. Its mean training reward also increases overall (Fig.~\ref{fig:propagation_reward}(b)). Compared with SAVN-CE, this corresponds to an 8.4-percentage-point increase in SR and an absolute SPL gain of 0.047, or relative improvements of 18.2\% and 11.5\%, respectively. Our method also reduces DTG from 5.810\,m to 4.176\,m compared with SAVN-CE, indicating that episodes end closer to the goal on average. These comparisons support the effectiveness of the proposed framework in unseen indoor scenes with predominantly indirect RF propagation. Figure~\ref{fig:qualitative_episode} illustrates four trajectories under varying RF propagation conditions.

\subsection{Ablation Study}
\label{sec:ablation}

Table~\ref{tab:ablation} examines directional uncertainty and RF history. Removing $\kappa$ from the goal token reduces SR from 0.546 to 0.489 and SPL from 0.456 to 0.420, supporting the value of explicitly encoding uncertainty. Using only current-step RF observations causes the largest drop, to 0.263 SR and 0.217 SPL, highlighting the importance of RF history.

\begin{table}[H]
    \caption{Ablation results on unseen scenes.}
    \label{tab:ablation}
    \centering
    \begin{tabularx}{\columnwidth}{@{}>{\raggedright\arraybackslash}X c c@{}}
        \toprule
        Variant & SR $\uparrow$ & SPL $\uparrow$ \\
        \midrule
        Full model & \textbf{0.546} & \textbf{0.456} \\
        w/o $\kappa$ in goal token & 0.489 & 0.420 \\
        Current-step RF only & 0.263 & 0.217 \\
        Deterministic goal (SAVN-CE) & 0.462 & 0.409 \\
        \bottomrule
    \end{tabularx}
\end{table}

Table~\ref{tab:exposure} reports the full model's performance under reduced RGB exposure.

\begin{table}[H]
    \caption{Navigation performance under reduced RGB exposure.}
    \label{tab:exposure}
    \centering
    \small
    \setlength{\tabcolsep}{4pt}
    \begin{tabular}{@{}c c c@{}}
        \toprule
        Exposure factor & SR $\uparrow$ & SPL $\uparrow$ \\
        \midrule
        1.0 & 0.546 & 0.456 \\
        0.7 & 0.515 & 0.436 \\
        0.5 & 0.511 & 0.436 \\
        \bottomrule
    \end{tabular}
\end{table}

Reducing the exposure factor from 1.0 to 0.7 lowers SR from 0.546 to 0.515 and SPL from 0.456 to 0.436. A further reduction to 0.5 yields an SR of 0.511, while SPL remains at 0.436 at the reported precision. Relative to normal exposure, the absolute decreases at an exposure factor of 0.5 are 0.035 in SR and 0.020 in SPL. These results suggest that the model tolerates the tested exposure reductions with modest performance degradation.

\section{Conclusion}
\label{sec:conclusion}

This paper presented a multimodal framework for indoor navigation toward an RF source. The Habitat--Sionna RT benchmark integrates embodied navigation with material-aware radio propagation simulation by assigning electromagnetic properties to semantically segmented MP3D meshes. It preserves detailed scene geometry and generates spatially aligned visual and RF observations as the robot moves. The proposed navigation method combines an uncertainty-aware RF goal representation with multimodal memory to address unreliable directional cues under multipath propagation. Experiments across 27 unseen scenes demonstrate improved navigation success and path efficiency over the evaluated baselines. Future work will focus on real-robot validation and larger antenna arrays for improved angular resolution.

\bibliographystyle{IEEEtran}
\bibliography{references}

\end{document}